# SNU_IDS at SemEval-2018 Task 12: Sentence Encoder with Contextualized Vectors for Argument Reasoning Comprehension

**Taeuk Kim, Jihun Choi** and **Sang-goo Lee**
Department of Computer Science and Engineering,
Seoul National University, Seoul, Korea
{taeuk,jhchoi,sglee}@europa.snu.ac.kr

## Abstract

We present a novel neural architecture for the Argument Reasoning Comprehension task of SemEval 2018. It is a simple neural network consisting of three parts, collectively judging whether the logic built on a set of given sentences (a claim, reason, and warrant) is plausible or not. The model utilizes contextualized word vectors pre-trained on large machine translation (MT) datasets as a form of transfer learning, which can help to mitigate the lack of training data. Quantitative analysis shows that simply leveraging LSTMs trained on MT datasets outperforms several baselines and non-transferred models, achieving accuracies of about 70% on the development set and about 60% on the test set.

## 1 Introduction

The Argument Reasoning Comprehension Task (Habernal et al., 2018) is a newly released task that tackles the core of reasoning in natural language argumentation, highlighting the importance of implicit warrants.

Even though the task could be regarded as simple binary classification, it is quite challenging in several perspectives. First, the task requires human-level reasoning to judge whether a claim supported by a reason and a warrant is logically correct. Second, common knowledge, which is not present in the input sentences themselves, is often required to solve the problem. Third, even though each instance of the data is helpful, the number of training data is relatively small to train prevailing complex neural models such as convolutional neural networks (Kim, 2014; Kalchbrenner et al., 2014) and recurrent neural networks (Hochreiter and Schmidhuber, 1997; Chung et al., 2014) with (or without) attention mechanisms (Liu et al., 2016; Lin et al., 2017).

In this paper, we propose a new architecture named **SECOVARC**[1] (Sentence Encoder with COnextualized Vectors for Argument Reasoning Comprehension) to deal with the complicated task. The main idea behind our model is that *transfer learning* can be a remedy to resolve the difficulties we face. With experimental results and analysis, we show that the simple neural model enhanced by transferred knowledge can be competitive, compared to complex models trained on the given data only.

## 2 Related Work

### 2.1 Argument Reasoning Comprehension

The argument reasoning comprehension task is a new dataset whose goal is to choose the correct implicit reasoning from two warrants, given a natural language argument with a reason and a claim. It consists of about 2K crowdsourced instances, each of which has a title and a short description of the debate from which the claim, reason, and two candidates arose. For more details, refer to Habernal et al. (2018).

### 2.2 Transfer Learning in NLP

Transfer learning is a classic technique in machine learning, which seeks to transfer beneficial knowledge from external resources to target models. It is well-known to be effective especially when one suffers from the lack of training data.

An important example showing the successfulness of transfer learning in natural language processing (NLP) is pre-trained word representations such as Word2Vec (Mikolov et al., 2013) and GloVe (Pennington et al., 2014), on which most of the modern models for NLP have been built.

---

[1] The implementation of our model is available at https://github.com/galsang/SemEval2018-task12

Furthermore, there are some recent works (Collobert et al., 2011; Mou et al., 2016; Min et al., 2017) that concentrate on pre-training more sophisticated neural modules over word embeddings, proving that transfer learning can be a key to boost the performance of NLP systems.

### 2.3 Unsupervised Sentence Representation

Following the success of unsupervised word representations, there arises another line of research to facilitate transfer learning in sentence-level. The idea is that a generic sentence encoder, which is pre-trained in an unsupervised way, can generate sentence representations suitable for downstream tasks.

For instance, Kiros et al. (2015) propose an approach called Skip-Thoughts vectors that abstracts the skip-gram of Word2Vec (Mikolov et al., 2013) to the sentence-level. Moreover, many other unsupervised methods (Le and Mikolov, 2014; Dai and Le, 2015; Hill et al., 2016; Gan et al., 2017; Chen, 2017) are also introduced as a way of building sentence representations.

### 2.4 Supervised Sentence Representation

Despite several attempts at learning sentence representations in an unsupervised manner, there has been no consensus established thus far, on which is the best method and can be adopted as a standard.

Meanwhile, sentence encoders trained on labeled datasets are proposed as an alternative, showing that they outperform the previous models even with the limited number of data. Conneau et al. (2017) suggest a method named InferSent, which uses a simple bidirectional LSTM (Long Short Term Memory, Hochreiter and Schmidhuber (1997)) with max-pooling trained on the Stanford Natural Language Inference (SNLI, Bowman et al. (2015)). And McCann et al. (2017) propose CoVe and demonstrate that the encoder part of the trained sequence-to-sequence (Sutskever et al., 2014) model for machine translation can be reused as a generic sentence encoder.

In the paper, we focus on supervised pre-training with external data as an instantiation of transfer learning.

## 3 Model

In this section, We describe SECOVARC (Figure 1) which takes a set of 3 sentences, i.e. a claim, reason, and warrant, as input and outputs a score between 0 and 1, indicating *how reasonable the claim is when it is based on the reason and the warrant*.

### 3.1 Model Design

Before jumping into the details, we explain about our motivation upon which the decisions on model design were made.

First, we let the model accept only one warrant instead of two candidates. This decision comes from the intuition that it may learn how to reason better when it *judges* whether the logic constructed on a set of a claim, reason, and warrant is plausible, instead of just *choosing* the more probable one between the two candidates.

Second, as mentioned earlier, one of the main concerns behind the model design is the lack of training data. To alleviate this problem, we decide to utilize transfer learning while maintaining the model as simple as possible (e.g. without introducing complex architectures such as attention mechanism).

### 3.2 Model Specification

In this part, we describe the details of the proposed model, which is composed of three layers.

#### 3.2.1 Encoding Layer

The encoding layer is the first part of our model, which is in charge of encoding three input sentences to corresponding sentence representations. In detail, it accepts the sequence of $n$ words $(w_1, w_2, \ldots, w_n)$ in a sentence at a time and outputs a fixed-length sentence representation $\mathbf{s}$. Note that the same generic encoder is used to encode each input sentence.

Formally, each one-hot encoded word $w_i \in \mathbb{R}^V$ of the input sentence is converted into the corresponding word vector $\mathbf{x}_i \in \mathbb{R}^{d_w}$ by a word embedding matrix $\mathbf{E} \in \mathbb{R}^{d_w \times V}$. Then, a sequence of the word vectors $\mathbf{x} = [\mathbf{x}_1, \mathbf{x}_2, \ldots, \mathbf{x}_n]$ is combined into $\mathbf{s} \in \mathbb{R}^{d_s}$ by an encoder. While a wide range of selection for the encoder is possible, in our case we utilize CoVe[2] (McCann et al., 2017) (with pooling operation), which is a two-layered Bi-LSTM pre-trained on large MT datasets, to obtain meaningful and contextualized sentence representations that would not be achieved if we train the encoder from scratch.

---
[2]Available at https://github.com/salesforce/cove

As a result, each representation for the claim ($\mathbf{s}_c$), reason ($\mathbf{s}_r$), and warrant ($\mathbf{s}_w$) is derived from $\mathbf{x}_c, \mathbf{x}_r$ and $\mathbf{x}_w$ as follows.

$$\mathbf{s}_c = \text{Pooling}(\text{CoVe}(\mathbf{x}_c))$$
$$\mathbf{s}_r = \text{Pooling}(\text{CoVe}(\mathbf{x}_r))$$
$$\mathbf{s}_w = \text{Pooling}(\text{CoVe}(\mathbf{x}_w))$$

From various options for the pooling operation, we use max-pooling, which selects the maximum value over each dimension of the output, and last-pooling that just selects the last state of the output. We call the CoVe encoder with max-pooling as SECOVARC-max and the encoder with last-pooling as SECOVARC-last.

### 3.2.2 Localization Layer

Although all of the input sentences (i.e. the claim, reason, and warrant) are encoded by the universal encoder, there is a need to make a difference among them so that each of the sentence representations keeps its own role. For this reason, the localization layer is introduced to project (or 'localize') each $\mathbf{s}$ onto its own semantic space.

We implement this layer simply in the form of three separate fully-connected layers, pursuing the intuition that our model should be simple. Therefore, a set of the sentence representations $\{\mathbf{s}_c, \mathbf{s}_r, \mathbf{s}_w\}$ is converted into $\{\mathbf{v}_c, \mathbf{v}_r, \mathbf{v}_w\} \in \mathbb{R}^{d_f}$ as follows.

$$\mathbf{v}_c = \tanh(\mathbf{W}_c \mathbf{s}_c + \mathbf{b}_c)$$
$$\mathbf{v}_r = \tanh(\mathbf{W}_r \mathbf{s}_r + \mathbf{b}_r)$$
$$\mathbf{v}_w = \tanh(\mathbf{W}_w \mathbf{s}_w + \mathbf{b}_w)$$

### 3.2.3 Output Layer

The output layer collects all features extracted from the previous layer and computes a final score between 0 and 1. To help the model make correct decisions, we introduce heuristic methods such as $|\mathbf{v}_w - \mathbf{v}_r - \mathbf{v}_c|$ and $\mathbf{v}_w \odot \mathbf{v}_r \odot \mathbf{v}_c$[3], inspired from the work of Mou et al. (2015) for the SNLI task.

In the end, a final feature $\mathbf{v}_f$ for computing a score $y \in \mathbb{R}$ ($0 \leq y \leq 1$) becomes a concatenation of the five vectors,

$$\mathbf{v}_f = \begin{bmatrix} \mathbf{v}_c \\ \mathbf{v}_r \\ \mathbf{v}_w \\ |\mathbf{v}_w - \mathbf{v}_r - \mathbf{v}_c| \\ \mathbf{v}_w \odot \mathbf{v}_r \odot \mathbf{v}_c \end{bmatrix}$$

[3] $\odot$: element-wise multiplication

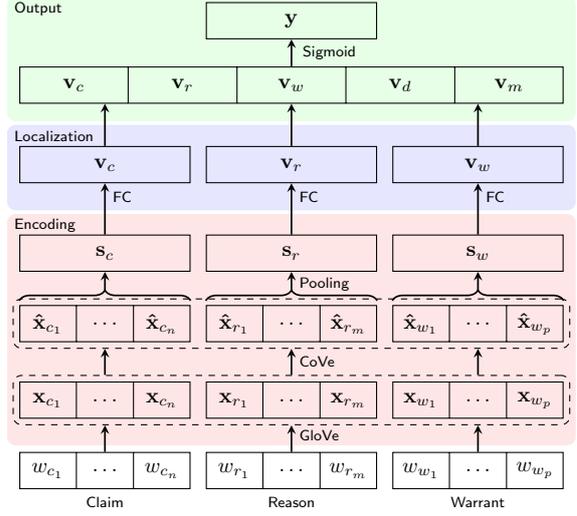

Figure 1: The architecture of SECOVARC. Dotted boxes represent the elements computed by parameter-shared modules (GloVe and CoVe) for all inputs. Note that $\mathbf{v}_d : |\mathbf{v}_w - \mathbf{v}_c - \mathbf{v}_r|$ and $\mathbf{v}_m : \mathbf{v}_w \odot \mathbf{v}_c \odot \mathbf{v}_r$. FC means a fully connected layer.

where $\mathbf{v}_f \in \mathbb{R}^{5d_f}$. Then, logistic regression (for simplicity) is performed on $\mathbf{v}_f$ to compute the final score.

$$y = \sigma(\mathbf{W}_f \mathbf{v}_f + \mathbf{b}_f)$$

During training, the score can be directly utilized to optimize the model. At test time, on the other hand, we derive $y_1$ and $y_2$ from the trained model with the input sentences such that

$$y_1 = \text{SECOVARC}(\mathbf{c}, \mathbf{r}, \mathbf{w}_1)$$
$$y_2 = \text{SECOVARC}(\mathbf{c}, \mathbf{r}, \mathbf{w}_2)$$

where $\mathbf{c}, \mathbf{r}, \mathbf{w}_1$ and $\mathbf{w}_2$ is the claim, the reason, the first warrant, and the second warrant respectively. Then, we select the warrant whose score is greater than that of the other as a final decision.

## 4 Experiment and Discussion

### 4.1 Data Manipulation

As our model requires only one warrant at a time, data preprocessing is inevitable before training. We manipulate the original data so that the correct warrant has a score of 1 and the opposite warrant has 0. Note that this pre-processing procedure has a side effect of doubling the original training data.

### 4.2 Training Details

The dimension of a word vector ($d_e$) is fixed to 300. And hyper-parameters for other vectors are set to $d_s = 600, d_f = 300$. We use 840B GloVe to initialize a word embedding matrix. Other model weights are randomly sampled from uniform distribution(-0.005, 0.005), except for the CoVe encoder, and biases are initialized with 0.

| Approach | Dev | (±) | Test | (±) |
|---|---|---|---|---|
| Human average | - | - | .798 | .162 |
| Human w/ training in reasoning | - | - | .909 | .114 |
| Random baseline | .473 | .039 | .491 | .031 |
| Language model | .617 | - | .500 | - |
| Attention | .488 | .006 | .513 | .012 |
| Attention w/ context | .502 | .031 | .512 | .014 |
| Intra-warrant attention | .638 | .024 | .556 | .016 |
| Intra-warrant attent. w/ context | .637 | .040 | .560 | .055 |
| SECOVARC (official record) | .731 | - | .565 | - |
| SECOVARC-last (w/o heuristics) | .701 | .011 | .559 | .019 |
| SECOVARC-last (w/ heuristics) | **.706** | .014 | .554 | .015 |
| SECOVARC-max (w/o heuristics) | .680 | .007 | .591 | .016 |
| SECOVARC-max (w/ heuristics) | .684 | .008 | **.592** | .016 |

Table 1: Comparison of baselines and variants of our model on the development set and the test set.

| Approach | Dev | (±) | Test | (±) |
|---|---|---|---|---|
| BoW | .677 | .006 | .502 | .014 |
| Bi-LSTM-last | .678 | .010 | .554 | .024 |
| Bi-LSTM-max | .670 | .011 | .543 | .027 |
| SECOVARC-last | **.706** | .014 | .554 | .015 |
| SECOVARC-max | .684 | .008 | **.592** | .016 |

Table 2: Experiment on the possibility of transfer learning in case of the argument reasoning comprehension task. Note that the heuristic methods are employed for all models.

Our model is trained using Adam (Kingma and Ba, 2014) optimizer with a learning rate 0.001 and a batch size 64. The maximum number of training epoch is limited to 10 and we choose the best model based on development accuracy. All parameters in the model, including the word vectors, are fine-tuned during training.

For regularization, L2-norm of the parameters is added to the Cross Entropy objective with the weight of 1e-5, and Dropout (Srivastava et al., 2014) technique is also applied with $p = 0.1$.

### 4.3 Experimental Results

Table 1 shows the accuracies of variants of our model and baselines (Habernal et al., 2018) on the development set and the test set. Due to the instability of results caused by random initialization, we report the mean and standard deviation of 20 experimental runs (with the same hyper-parameters) for each model.

The reported results show that SECOVARC-last (w/ heuristics) outperforms all the baselines on the development set, with a mean accuracy of 70.6%. However, it is SECOVARC-max (w/ heuristics) that performs best on the test set, with a mean accuracy of 59.2%. We submitted an instance obtained from SECOVARC-last (w/ heuristics) and achieved the official result of 56.5% on the leaderboard. Table 1 also demonstrates that our model benefits from the heuristics applied in the output layer, except for the test accuracy of SECOVARC-last.

#### 4.3.1 Does transfer learning really work?

Even with the promising outcome presented by SECOVARC, an issue remains regarding how to show the effectiveness of transfer learning for the task. For this objective, we conduct additional experiments with three baselines called BoW, Bi-LSTM-last, and Bi-LSTM-max. Bi-LSTM-last and Bi-LSTM-max have the same architecture with SECOVARC, but the Bi-LSTMs in the encoding layer are randomly initialized rather than pre-trained. BoW is different from our proposed model in that it leverages the average of word vectors as a sentence representation instead of using CoVe with pooling.

Table 2 reports the comparison of the baselines and the variants of our model. The results show that our model consistently outperforms the baselines which are trained from scratch. Moreover, the smaller deviations of SECOVARCs demonstrate that transfer learning can lead to more stable and successful training of models.

## 5 Conclusion

In this paper, we present a novel neural architecture called SECOVARC, that utilizes a two-layered Bi-LSTM trained first on a large amount of machine translation data. And we demonstrate that the neural model for the argument reasoning comprehension task can benefit from transfer learning when it is properly designed.

As a future work, there is a way to apply contemporary works for generic sentence encoders such as Subramanian et al. (2018) and Peters et al. (2018) instead of CoVe. On the other hand, we can consider expanding the data itself directly with sophisticated rules or heuristics.

## Acknowledgments

This work was supported by the National Research Foundation of Korea (NRF) grant funded by the Korea government (MSIT) (NRF-2016M3C4A7952587).